\documentclass[10pt,twocolumn,letterpaper]{article}

\usepackage{wacv}
\usepackage{times}
\usepackage{epsfig}
\usepackage{graphicx}
\usepackage{amsmath}
\usepackage{amssymb}
\usepackage{amsfonts}
\usepackage{makecell}
\usepackage{pifont}
\usepackage{algorithm}
\usepackage{algorithmicx}
\usepackage{algpseudocode}

\def\wacvPaperID{****} 

\wacvfinalcopy 

\def\assignedStartPage{1} 

\ifwacvfinal
\usepackage[breaklinks=true,bookmarks=false]{hyperref}
\else
\usepackage[pagebackref=true,breaklinks=true,colorlinks,bookmarks=false]{hyperref}
\fi

\ifwacvfinal
\else
\fi

\begin{document}

\title{Dual-Student Knowledge Distillation Networks for Unsupervised Anomaly Detection}

\author{Liyi Yao\\
University of Southern California\\
Los Angeles, California, USA\\
{\tt liyiyao@usc.edu}
\and
Shaobing Gao\\
Sichuan University\\
Chengdu, Sichuan, China\\
{\tt gaoshaobing@scu.edu.cn}
}

\maketitle

\begin{abstract}
Due to the data imbalance and the diversity of defects, student-teacher networks (S-T) are favored in unsupervised anomaly detection, which explores the discrepancy in feature representation derived from the knowledge distillation process to recognize anomalies.
However, vanilla S-T network is not stable. Employing identical structures to construct the S-T network may weaken the representative discrepancy on anomalies. But using different structures can increase the likelihood of divergent performance on normal data.
To address this problem, we propose a novel dual-student knowledge distillation (DSKD) architecture. Different from other S-T networks, we use two student networks a single pre-trained teacher network, where the students have the same scale but inverted structures.
This framework can enhance the distillation effect to improve the consistency in recognition of normal data, and simultaneously introduce diversity for anomaly representation.
To explore high-dimensional semantic information to capture anomaly clues, we employ two strategies. First, a pyramid matching mode is used to perform knowledge distillation on multi-scale feature maps in the intermediate layers of networks. Second, an interaction is facilitated between the two student networks through a deep feature embedding module, which is inspired by real-world group discussions.
In terms of classification, we obtain pixel-wise anomaly segmentation maps by measuring the discrepancy between the output feature maps of the teacher and student networks, from which an anomaly score is computed for sample-wise determination.
We evaluate DSKD on three benchmark datasets and probe the effects of internal modules through ablation experiments.
The results demonstrate that DSKD can achieve exceptional performance on small models like ResNet18 and effectively improve vanilla S-T networks.
\end{abstract}

\section{Introduction}
\label{sec:introduction}
\begin{figure}[!t]
\centering
\includegraphics[width=\columnwidth]{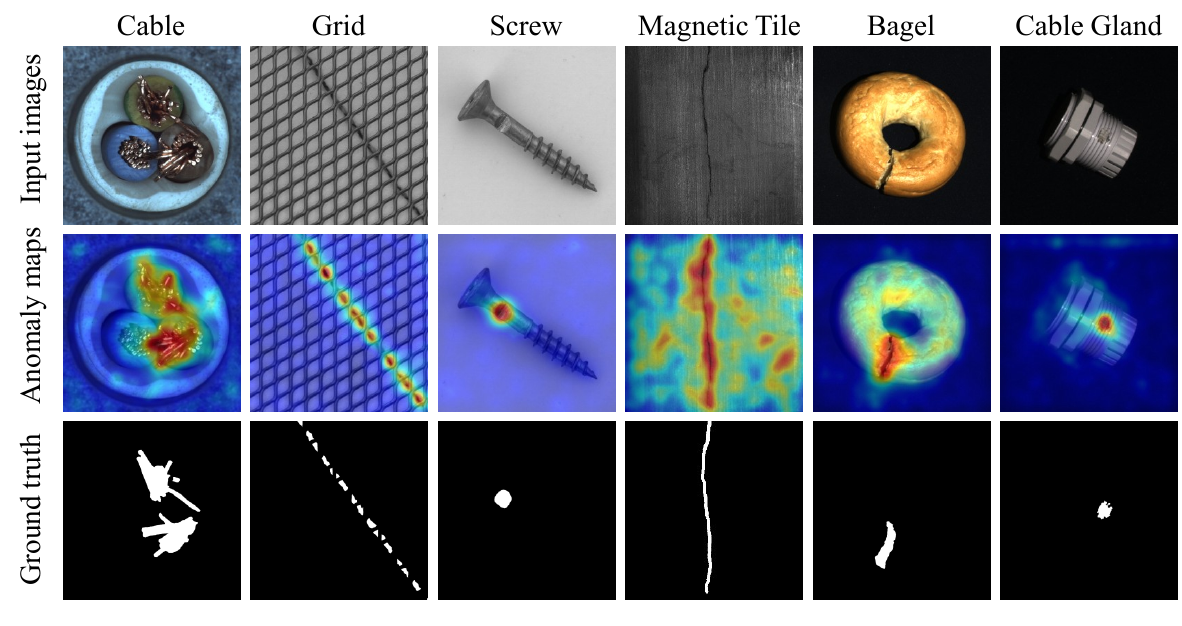}
\caption{Examples of anomaly localization or segmentation. From the first to the third line, anomaly samples, anomaly maps generated by our proposed model, and ground truth are shown, respectively.}
\label{example}
\end{figure}

In industrial manufacturing, automated \textbf{Anomaly detection} (AD) usually plays a significant role in quality control, which refers to recognize the defects and faults based on vision technology \cite{tao2022survey}.
Restrictively speaking, anomaly detection determines whether there are anomalies at the image level. In terms of pixels, the goal is to judge whether a pixel falls into the anomalous area, which is also known as \textbf{Anomaly Localization} (AL) \cite{wan2022pfm}. Fig. \ref{example} shows pixel-wise examples of AL.
To remove ambiguity, we explicitly point out the image-level anomaly detection and the pixel-level anomaly localization in the Sec. \ref{sec:experiments}. 
And for the sake of simplicity, in other sections, we refer to AD and AL collectively as ``anomaly detection''.

In general, there are two main challenges for AD.
First, defects are usually rare in the normal manufacturing process. It is difficult to acquire as many anomalous samples as anomaly-free samples, which results in data imbalance. And since most current vision recognition algorithms are data-drive, this drawback can damage their performance \cite{wan2022pfm}.
On the other hand, the anomalous patterns are diverse and datasets for training can not contain all types of anomalies. Defects probably exist in the texture of the surface or in the inner structure, and their forms can be stains, wear, scratches, the absence of some parts, or unknown.
Therefore, without anomaly reference, supervised learning can hardly achieve accurate detection, while unsupervised learning methods can be a better choice.

\begin{figure}[!t]
\centering
\includegraphics[width=\columnwidth]{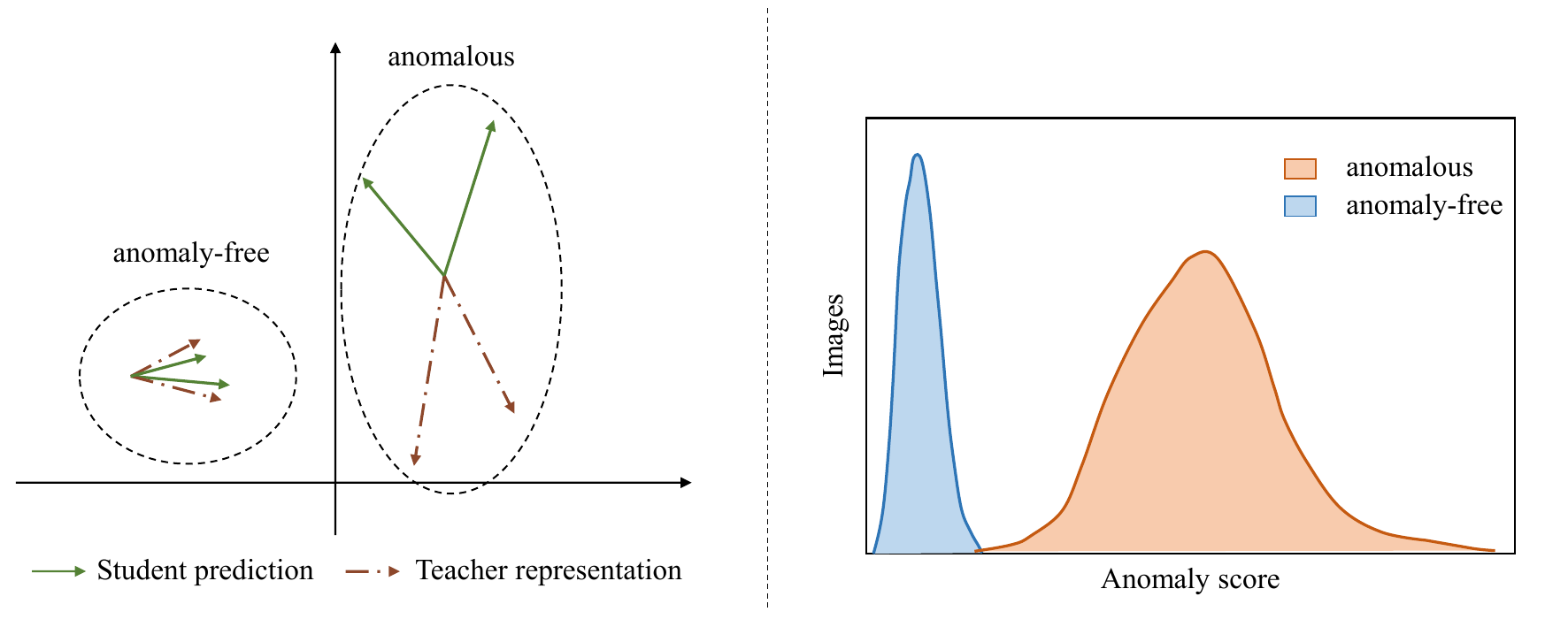}
\caption{Principles of knowledge distillation in anomaly detection. In the left figure, the student and the teacher have similar representations of anomaly-free patterns but differ significantly in anomalies. Given this, we can calculate the anomaly scores in the right figure.}
\label{principle}
\end{figure}

\begin{figure*}[!t]
	\centering
	\includegraphics[width=2\columnwidth]{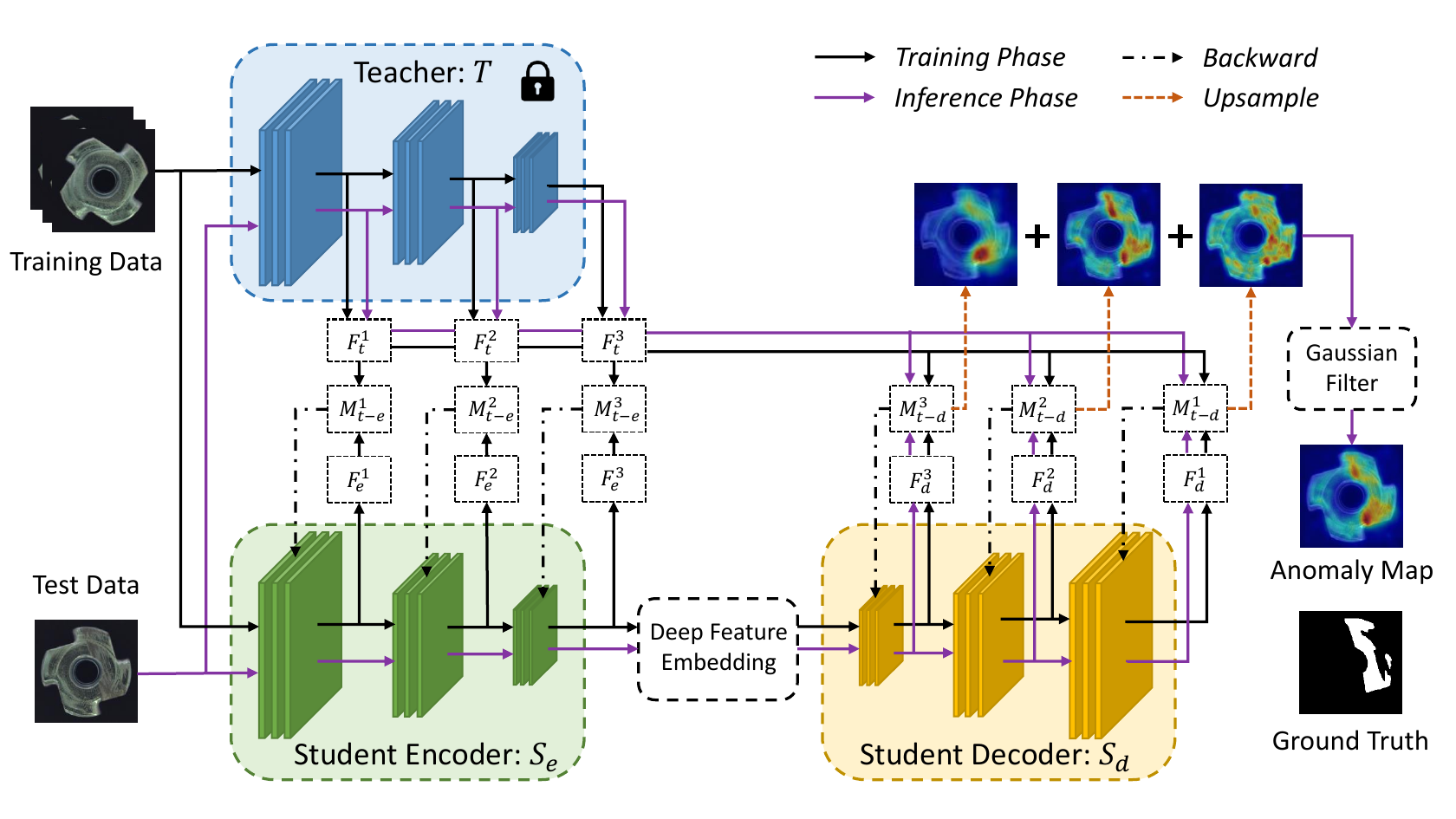}
	\caption{The framework of the proposed dual-student knowledge distillation (DSKD) model.}
	\label{architecture}
\end{figure*}

To tackle these problems, some traditional vision technologies are applied in AD for feature extraction and classification, like one-class support vector machine (ocsvm) \cite{scholkopf1999ocsvm}.
In terms of pattern recognition, deep learning-based approaches show great success in AD. And these approaches can be broadly classified into two categories, \textbf{reconstruction-based} models \cite{baur2019deep, bergmann2018improving, rudWeh2022csflow, rudolph2020differnet, schlegl2017anogan, Schlegl2019fanogan} and \textbf{deep feature embedding-based} models \cite{cohen2021spade, defard2020padim, roth2022patchcore, dehaene2020favae, bergmann2020uninformed, wang2021student}.
The former learn how to reconstruct data in defect-free distribution, and capture the discrepancy between the source and reconstructed data. In concrete, if input images are non-anomalous, the trained model can generate images in similar data distribution. Otherwise, the generated images show evident difference from the inputs \cite{schlegl2017anogan}.
With respect to the feature embedding-based methods, they leverage the extracted feature information to estimate the anomalous patterns, usually in the latent space. For unsupervised learning, the feature extraction process is usually based on pre-trained models \cite{wan2022pfm}.

Currently, many studies use the \textbf{knowledge distillation} (KD) paradigm \cite{hinton2015distilling} for anomaly detection, which falls into the feature embedding-based category \cite{bergmann2020uninformed, salehi2020multiresolution, wang2021student, deng2022reverse, dehaene2020favae}.
KD is first proposed to transfer knowledge from a pre-trained large, complex model (teacher) to smaller, simpler models (student) for model compression \cite{hinton2015distilling}. A random initialized model is like a naive child, \emph{i.e.} student, who needs to learn and understand the world, and a pre-trained model is an expert, \emph{i.e.} teacher, with rich knowledge. The new model can learn from the output of the pre-trained model rather than ground truth labels. Then, parameter-wise information and knowledge stored in the teacher network can be transferred to the student, which is called \textbf{distillation}.
Since both the teacher and the student are involved in this process, it is also known as \textbf{student-teacher (S-T)} networks. 

In anomaly detection, the teacher network is pre-trained on some large-scale datasets, like ImageNet, and its parameters are frozen in the training process.
In the meanwhile, the student is only fed with the anomaly-free data and the distilled knowledge, \emph{i.e.} feature maps, transferred from the teacher. In other word, features extracted by the teacher are embedded into the student to help the student learn how to represent features of normal data.
Therefore, the teacher and the trained student can represent anomaly-free images in the same way, but show discrepancy in the representation of anomalies, as shown in Fig. \ref{principle}.
By analogy, if a student only learn math from a well-educated teacher, they are likely to give different answers to some geological questions.
Then we can measure the distance between the output feature maps of the teacher and the student to estimate the anomalous patterns by calculating the anomaly score, where a higher score means a higher probability of anomalies. 
If the anomaly score is higher than the threshold, it can be considered as anomaly as exhibited in \eqref{classification}. In our work, we perform normalization on the anomaly score and set the threshold to $0.5$.
\begin{equation}
A(score)=\left\{
\begin{aligned}
&True, &score\geqslant threshold,\\
&False, &score < threshold.\\
\end{aligned}
\right.
\label{classification}
\end{equation}

To exploit the such representation discrepancy between the teacher and the student, some studies \emph{et al.} \cite{salehi2020multiresolution} use asymmetric S-T networks, where the teacher network usually has a larger and more complex backbone like the classic knowledge distillation paradigm.
However, this architecture can also expose the representation discrepancy in anomaly-free data sometimes, which damages the detection accuracy.
In contrast, some other works \cite{wang2021student} use a symmetric structure, where both teacher and student are based on the same backbones. 
The limitation of such architecture is that the teacher and the student probably perform similarly on anomalous data, which also makes negative effects as well.
An ideal architecture should highlight differences in anomaly representation while maintaining representative similarity for anomaly-free data.

For this reason, we proposed a \textbf{dual-student knowledge distillation paradigm (DSKD)} for AD, as shown in Fig. \ref{architecture}.
In DSKD, we add an extra student network and there are totally three players involved, a teacher $T$ and two students $S_e$ and $S_d$. $T$ and $S_e$ are based on ResNet18 \cite{he2016resnet}, while the backbone of $S_d$ is completely reversed from $S_e$.
In this way, the symmetric combination of $T$ and $S_e$ helps produce similar results with normal samples and the inverted structure can strengthen the representation discrepancy on anomalous samples.
We regard the methods of Salehi \emph{et al.} \cite{salehi2020multiresolution} and Wang \emph{et al.} \cite{wang2021student} as baselines for the following experimental study.

In DSKD, $T$ has been pre-trained on ImageNet-1K, and its parameters are frozen.
The first student $S_e$ works as an encoder to extract features from the input. And the second student $S_d$ is a decoder, whose function is to decode information from the high-dimensional features that extracted by $S_e$.
The distillation processes between two students and the teacher is performed separately and independently.
Since feature maps of different layers can focus on different types of features in hierarchical networks \cite{salehi2020multiresolution}, DSKD fuses multi-scale intermediate feature maps to explore the high-dimensional semantic information to enhance distillation effects rather than simply measuring the final outputs. 
The multi-scale feature fusion exists not only in the distillation between the teacher and student networks, but also in the collaboration between student networks.
$S_e$ and $S_d$ are connected by a bottleneck block, which fuses the intermediate feature maps of $S_e$ through several convolutional blocks and embeds them to $S_d$ in format of low-dimensional vectors.
This is inspired by real-world education scenarios, where collaboration among students can aid in a better understanding of the knowledge imparted by the teacher. 
Regarding the reasoning phase, \emph{i.e.} anomaly inference, data processed by $S_e$ and $S_d$ sequentially is matched with the outputs of $T$ to obtain anomaly maps.
The technical details are discussed in Sec. \ref{sec:proposed method}. 
And in Sec. \ref{sec:experiments}, we discuss the experiments and analyze the results.

Our contributions can be summarized as follows:

\begin{enumerate}
\item We propose a novel dual-student knowledge distillation paradigm consisting of a teacher network and two student networks for unsupervised anomaly detection, where the networks have the same scale but the two students have totally inverted structures This method can effectively improve the distillation effects by enhancing the diversity of anomalous feature representations and reduce the discrepancy on anomaly-free data.
\item We design the multi-scale feature fusion block to explore high-dimensional semantic information, which matches intermediate feature maps between the teacher and student networks based on a feature pyramid instead of only using the final outputs. Intermediate anomaly maps are upsampled to the same size and summed together for anomaly inference.
\item The two student networks are connected by a bottleneck design, where the deep features of the first student are embedded. This deep feature embedding module uses convolutional blocks to downsample multi-scale feature maps and do channel-wise feature fusion. The collaboration of two student networks can also benefit the detection results.
\item We performed experiments on three benchmark datasets, MVTec AD \cite{bergmann2021mvtec}, MVTec 3D-AD \cite{bergmann2021mvtec3d}, and Magnetic Tile Defect \cite{huang2020mt}. Our proposed method outperforms baselines and achieve good results. And the ablation experiments prove the effectiveness of the dual-student architecture and its interior modules.
\end{enumerate}

\begin{algorithm}
\caption{DSKD Training Procedure} \label{training}
\textbf{Input:} Training dataset $\mathcal{I}^t=\{I_1^t, I_2^t, ..., I_m^t\}$, parameters $\theta_t$ of $T$, epoch number $n$; \\
\textbf{Output:} Parameters $\theta_e$ of $S_e$, parameter $\theta_d$ of $S_d$;
\begin{algorithmic}[1]
\State randomly initialize parameters of $\theta_e$ and $\theta_d$;
\State $T \gets T.\text{load\_weights}(\theta_t)$;
\For{$i \gets 1$ to $n$}
    \For {$j \gets 1$ to $m$}
        \State $S_e \gets S_e.\text{load\_weights}(\theta_e)$;
        \State $S_d \gets S_d.\text{load\_weights}(\theta_d)$;
        \State $\hat{F}_t \gets  {\ell}_2\_\text{normalization}(T(I_j^t))$;
        \State $\hat{F}_e \gets  {\ell}_2\_\text{normalization}(S_e(I_j^t))$;
        \State $\hat{F}_d \gets  {\ell}_2\_\text{normalization}(S_d(F_{emb}))$;
        \For {$k \gets 1$ to $\text{length}(T)$}
            \State \parbox[t]{\dimexpr 0.8\linewidth-\algorithmicindent}{Obtain anomaly map $M_{t-e}^k$ between $T$ and $S_e$ based on Eq. \eqref{ano_map};}
            \State \parbox[t]{\dimexpr 0.8\linewidth-\algorithmicindent}{Obtain anomaly map $M_{t-d}^k$ between $T$ and $S_d$ based on Eq. \eqref{ano_map};}
        \EndFor
        \State $\ell_{e} \gets \text{mean}(M_{t-e})$;
        \State $\ell_{d} \gets \text{mean}(M_{t-d})$;
        \State Update parameters $\theta_{e}$;
        \State Update parameters $\theta_{d}$;
    \EndFor
\EndFor
\end{algorithmic}
\end{algorithm}

\section{Related Works}
\label{sec:related works}
Anomaly detection is first considered a kind of one class classification (OCC) task where some traditional machine learning and vision methods are employed.
Due to the diversity of anomalous features, data-driven deep learning technologies become more popular, which can be categorized into reconstructed-based methods and deep feature embedding-based methods.

\subsection{Classical Methods}
Classical AD methods generally process vectors in a high-dimensional space. The one-class support vector machine (OCSVM) \cite{scholkopf1999ocsvm} maximized the distance between the hyperplane and the origin in the feature space and estimated the probability density area for one-class classification. 
Support vector data description (SVDD) \cite{tax2004svdd} uses a hypersphere instead of a hyperplane, which yields better results in defects detection. 
Deep-SVDD \cite{hu2021semantic} maps data in a smaller hypersphere by deep networks. And Patch-SVDD \cite{yi2020patch} implements one-class detection at the pixel level.
However, these classical methods usually have weak performance in generalization and are inefficient on large-scale datasets.

\subsection{Reconstruction-based Methods}
These methods are inspired by the idea that models trained with anomaly-free data can only reconstruct both anomalous and normal data in anomaly-free distribution. Therefore, the divergence between the input and output data distributions can be used to identify anomalous patterns \cite{tao2022survey}. 
Vanilla auto-encoder (Vanilla AE) \cite{baur2019deep} is applied in anomaly segmentation for brain images. And Bergmann \emph{et al.} \cite{bergmann2018improving} use the structural similarity (SSIM) metric as the optimization objective function.

Additionally, GAN-based models also have good abilities in feature reconstruction. Schlegl \emph{et al.} \cite{schlegl2017anogan} propose Anomaly GAN (AnoGAN), which uses the Wasserstein distance to measure the discrepancy between normal and anomalous data. And fast AnoGAN (f-AnoGAN) \cite{Schlegl2019fanogan} optimizes the architecture by using a trained decoder as the generator and requires less computation resources.

Unlike other generative models, normalizing flows (NF) \cite{rezende2016nf} has advantages in density estimation. Rudolph \emph{et al.} \cite{rudolph2020differnet} propose an NF-based model, DifferNet, for anomaly detection. To localize anomalous regions, fully convolutional cross-scale normalizing (CS-Flow) \cite{rudWeh2022csflow} retains the spatial arrangement mode in the latent space and can process multi-scale feature maps. 

\subsection{Deep Feature Embedding-based Methods}
Since errors in reconstruction can result in mistakes in detection \cite{tao2022survey}, some studies identify anomalies with extracted features mapped to low-dimension representation, known as \textbf{embedding} \cite{golinko2019embedding}. And the representative discrepancy can be measured in the embedding space rather than reconstruction.
Considering that pre-trained models retain high performance of capturing deep semantic information in downstream tasks, feature extractors are generally pre-trained on large-scale datasets like ImageNet. 
Cohen \emph{et al.} \cite{cohen2021spade} make improvements on deep pre-trained k-nearest neighbor methods via a feature pyramid. 
Defard \emph{et al.} \cite{defard2020padim} propose a patch distribution modeling (PaDiM) framework that implements patch embedding with a pre-trained CNN model and multivariate Gaussian distributions. 
Inspired by this patch-level feature processing method, Roth \emph{et al.} \cite{roth2022patchcore} present the PatchCore model, which deposits extracted features in a compact memory bank and uses the nearest neighbor search algorithm to simplify the computation in the inference period.

In terms of knowledge distillation for AD, the teacher is supposed to impart only knowledge about anomaly-free data to the student so that they can perform differently on anomalous data.
Bergmann \emph{et al.} \cite{wang2021student} introduce KD in anomaly detection, where the teacher network is pre-trained with its parameters frozen, and the student only learns the non-anomalous distribution.
S-T networks can directly capture defective information at the pixel level by comparing the output feature maps of the teacher and the student networks. 
To explore high-dimensional information of deep layers, Salehi \emph{et al.} \cite{salehi2020multiresolution} propose a multi-resolution KD (MKD) model, which leverages the features in intermediate layers. And MKD employs an asymmetric architecture
On the contrary, Wang \emph{et al.} \cite{wang2021student} fused the outputs of intermediate layers into a feature pyramid within symmetric S-T networks.
Furthermore, some studies use feature reconstruction strategies to improve the quality of distillation.
Dehaene \emph{et al.} proposed a feature-augmented Variational Auto-Encoder (FAVAE) \cite{dehaene2020favae} framework consisting of a feature extractor and a VAE module. The extracted features can be embedded in the decoder as a S-T network.
Based on auto-encoder, a reverse distillation \cite{deng2022reverse} approach uses an encoder as the teacher and a decoder as the student to leverage the teacher's embedding information.
In our dual-student S-T networks, there are three players employed instead of only one pair of teacher and student networks.
Two inverted student networks learn from the same teacher in the encoding and decoding aspects, respectively. In this way, it can amplify the discrepancy between the representation of anomalous features of the teacher and the student and alleviate that of anomaly-free features, as discussed in Sec. \ref{sec:introduction}.

\begin{figure}[!t]
\centering
\includegraphics[width=\columnwidth]{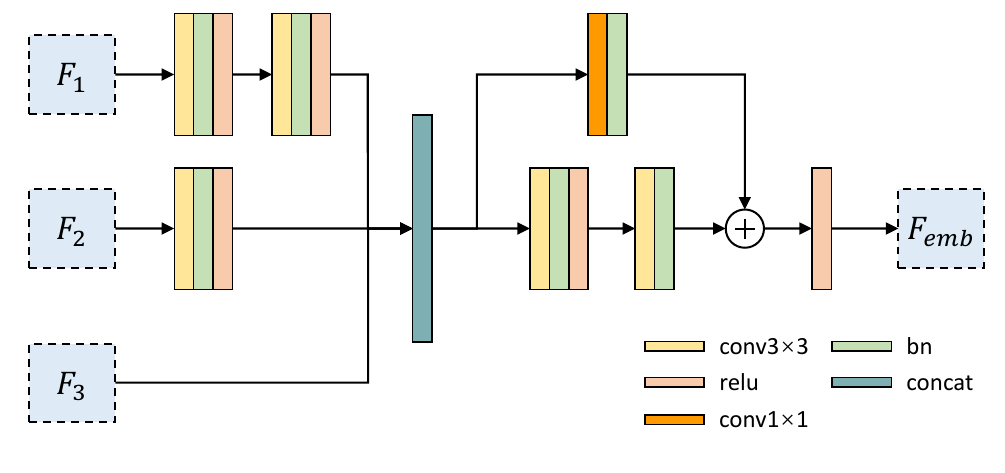}
\caption{Deep feature embedding process. Feature maps from different layers are resized to the same scale and then are downsampled by convolutional modules. The embedding carries rich semantic information from different intermediate layers.}
\label{fig4}
\end{figure}

\section{Proposed Method}
\label{sec:proposed method}
In this section, we present the our model in detail and discuss its technical implementation. Fig. \ref{architecture} exhibits the overview of our proposed framework. For better expressions, the training dataset and test dataset are respectively marked as $\mathcal{I}^t=\{I_1^t, I_2^t, ..., I_m^t\}$ and $\mathcal{I}^a=\{I_1^a, I_2^a, ..., I_m^a\}$. $\mathcal{I}^t$ only consists of anomaly-free samples, while $\mathcal{I}^a$ has anomaly-free and anomalous samples. And the $m$th image is $I_m \in \mathbb{R}^{w\times h\times c}$, where $w$ and $h$ represent the width and height of the image, and $c$ denote the number of channels.

\subsection{Dual-Student Knowledge Distillation}
The classical KD method employs one pair of S-T networks for knowledge transfer and model size compress. Given a pre-trained teacher network $T$ and a target student network $S$, the objective can be shown as:
\begin{equation}\label{kd}
\begin{split}
    { \underset{\theta}{{\arg\min} \, {\ell}_{KD}}} = \mathbf{D}(T_{\theta^*}(x), S_{\theta}(x)) + \lambda\mathbf{D}(y, S_{\theta}(x)),
\end{split}
\end{equation}
where $\theta^*$ and $\theta$ denote the parameters of $T$ and $S$, $y$ is the ground truth, and $\mathbf{D}$ represents a certain distance metric. Kullback-Leible (KL) divergence is used to measure the distance, which is proven to be equivalent to the Euclidean distance, \emph{i.e.} ${\ell}_2$-distance, in the optimization process according to the mild assumption \cite{hinton2015distilling}.

Unlike other S-T network-based approaches, we add a new student to enhance inconsistencies in the feature representation of out-of-distribution data.
As shown in Fig. \ref{architecture}, DSKD has three parts, teacher $T$, students $S_e$ and $S_d$. Their feature maps can be represented as $F_t^k(I_m), F_e^k(I_m), F_d^k(I_m) \in \mathbb{R}^{w_k\times h_k\times c_k}$, where $t$, $e$, $d$, respectively, denote $T$, $S_e$, $S_d$, and $k$ is the index of the intermediate layer.
We hypothesize that the student $S_e$ can progressively acquire knowledge related to anomaly-free features by learning from $T$, in turn, help $S_d$ enhance its understanding of the same knowledge transferred from $T$.
Because in a hierarchical network different intermediate layers probably focus on different semantic information \cite{salehi2020multiresolution}, we integrate the intermediate feature maps like a feature pyramid rather than only using the last layer's outputs.
Specifically, in our work, we select the first three blocks in ResNet18, \emph{i.e.} conv2\_x, conv3\_x, conv4\_x \cite{he2016resnet}, so the total number of intermediate layers $K=3$.
The vectors on each feature map are localized by the indices $i$ and $j$, and noted as $F_t^k(I_m)_{ij}, F_e^k(I_m)_{ij}, F_d^k(I_m)_{ij} \in \mathbb{R}^{c_k}$.
These vectors are normalized by ${\ell}_2$-distance as follows:
\begin{equation}\label{l2_norm}
\begin{split}
    {\hat{F}_t^k(I_m)}_{ij}=
    \frac{{F_t^k(I_m)}_{ij}}{{||{F_t^k(I_m)}_{ij}||}_{{\ell}_2}^2},&\quad
    {\hat{F}_e^k(I_m)}_{ij}=
    \frac{{F_e^k(I_m)}_{ij}}{{||{F_e^k(I_m)}_{ij}||}_{{\ell}_2}^2},\\
    {\hat{F}_d^k(I_m)}_{ij}=
    \frac{{F_d^k(I_m)}_{ij}}{{||{F_d^k(I_m)}_{ij}||}_{{\ell}_2}^2}.
\end{split}
\end{equation}

$T$ is pre-trained on a large-scale dataset, ImageNet-1K, which is regarded as prior knowledge. 
In the training phase, $T$'s parameters are frozen and $T$ is used to extract features from data, providing reference for students.
$S_e$ has the same network as $T$, learning from $T$ in the distillation process.
$S_d$ is reversed from $S_e$ with a completely transposed backbone, where downsampling operations are replaced by upsampling operations.
The two students networks are connected by a deep feature embedding block, which is discussed in Sec. \ref{subsec:embedding}. 
$S_d$ decodes the embedded data flow and upsamples the low-dimensional vectors to the same size as the input images. 
Regarding the knowledge distillation, we measure the pixel-wise distance of each intermediate feature map between the teacher and student networks and obtain the multi-scale anomaly maps.
To match the resolutions, the distillation is performed in intermediate layers with the same sizes, as exhibited in Fig. \ref{architecture}.
$T$ transfers the knowledge to $S_e$ and $S_d$ separately, and $S_d$ also receives the knowledge from $S_e$.
Algo. \ref{training} describes the training procedure.

To evaluate the discrepancy between the teacher and the students, DSKD uses a combined loss function. The first part is ${\ell}_2$-distance, minimizing the Euclidean distance. $t-e$ and $t-d$ separately denote the distillation between $T$ and $S_e$, and between $T$ and $S_d$, as shown in Eq. \eqref{l2_loss}.

\begin{equation}\label{l2_loss}
\begin{split}
    {{\ell}_2^{t-e}(I_m)}_{ij}=\frac{1}{2}{||{\hat{F}_t^k(I_m)}_{ij}-{\hat{F}_e^k(I_m)}_{ij}||}_{{\ell}_2}^2,\\
    {{\ell}_2^{t-d}(I_m)}_{ij}=\frac{1}{2}{||{\hat{F}_t^k(I_m)}_{ij}-{\hat{F}_d^k(I_m)}_{ij}||}_{{\ell}_2}^2,
\end{split}
\end{equation}

The other part is based on cosine similarity, which measures the directional distance between two vectors.
To minimize the loss function for optimization, this part ${\ell}_{cos}$ is written as follows:

\begin{equation}\label{cos_sim}
\begin{split}
    {{\ell}_{cos}^{t-e}(I_m)}_{ij}=1-\frac{{({\hat{F}_t^k(I_m)}_{ij})}^T\cdot {\hat{F}_e^k(I_m)}_{ij}}{{||{\hat{F}_t^k(I_m)}_{ij}||\enspace||{\hat{F}_e^k(I_m)}_{ij}||}},\\
    {{\ell}_{cos}^{t-d}(I_m)}_{ij}=1-\frac{{({\hat{F}_t^k(I_m)}_{ij})}^T\cdot {\hat{F}_d^k(I_m)}_{ij}}{{||{\hat{F}_t^k(I_m)}_{ij}||\enspace||{\hat{F}_d^k(I_m)}_{ij}||}}.
\end{split}
\end{equation}

The value of $(i, j)$ pixel on the anomaly map of the $k$th layer can be obtained by Eq. \eqref{ano_map}, where $\lambda$ is a coefficient.
To integrate anomaly maps in different sizes, we sum up pixels of all anomaly maps, and take the mean values as the total loss as described in Eq. \ref{loss} where $\ell_{e}$ denotes the loss of $S_e$, and $\ell_{d}$ denotes that of $S_d$.

\begin{equation}\label{ano_map}
\begin{split}
    {M_{t-e}^k(I_m)}_{ij}=\lambda {{\ell}_2^{t-e}(I_m)}_{ij}+
    {{\ell}_{cos}^{t-e}(I_m)}_{ij},\\
    {M_{t-d}^k(I_m)}_{ij}=\lambda {{\ell}_2^{t-d}(I_m)}_{ij}+
    {{\ell}_{cos}^{t-d}(I_m)}_{ij},
\end{split}
\end{equation}

\begin{equation}\label{loss}
\begin{split}
    \ell_{e}(I_m)=\frac{1}{K}\sum_{k=1}^{K}[\frac{1}{h_kw_k}\sum_{i=1}^{h_k}\sum_{j=1}^{w_k}
    {M_{t-e}^k(I_m)}_{ij}],\\
    \ell_{d}(I_m)=\frac{1}{K}\sum_{k=1}^{K}[\frac{1}{h_kw_k}\sum_{i=1}^{h_k}\sum_{j=1}^{w_k}
    {M_{t-d}^k(I_m)}_{ij}].
\end{split}
\end{equation}

\begin{algorithm}
\caption{DSKD Anomaly Inference} \label{inference}
\textbf{Input:} Test image $I^a$, parameters $\theta_t$ of $T$, parameters $\theta_e$ of $S_e$, parameters $\theta_d$ of $S_d$;\\
\textbf{Output:} Anomaly maps $\widetilde{M}$, detection result $A$
\begin{algorithmic}[1]
\State $M \gets \text{zeros}(I^a.\text{height}, I^a.\text{width})$;
\State $T \gets T.\text{load\_weights}(\theta_t)$;
\State $S_e \gets S_e.\text{load\_weights}(\theta_e)$;
\State $S_d \gets S_d.\text{load\_weights}(\theta_d)$;
\State $\hat{F}_t \gets  {\ell}_2\_\text{normalization}(T(I_a))$;
\State $\hat{F}_e \gets  {\ell}_2\_\text{normalization}(S_e(I_a))$;
\State $F_{emb} \gets \text{feature\_embedding}(\hat{F}_e)$;
\State $\hat{F}_d \gets  {\ell}_2\_\text{Normalization}(S_d(F_{emb}))$;
\For {$k \gets 1$ to $\text{length}(T)$}
    \State \parbox[t]{\dimexpr 0.8\linewidth-\algorithmicindent}{Obtain anomaly map $M_{t-d}^k$ between $T$ and $S_d$ based on Eq. \eqref{ano_map};}
    \State $M \gets M + \text{upsample}(M_{t-d}^k)$;
\EndFor
\State Denoise following Eq. \eqref{denoise}: $\widetilde{M}=G_{\sigma}(M, ~\sigma=4.0)$;
\State Obtain the anomaly score: $score=\text{max}(\widetilde{M})$;
\State $threshold \gets 0.5$;
\If {$score \geqslant threshold$}
    \State $A \gets \text{True}$;
\Else
    \State $A \gets \text{False}$
\EndIf
\end{algorithmic}
\end{algorithm}

\subsection{Deep Feature Embedding}
\label{subsec:embedding}
DSKD uses the deep feature embedding (DFE) module as a bottleneck to connect of $S_e$ and $S_d$ and activate their collaboration.
$S_e$ learns to encode anomaly-free data from $T$ and transfers one-way data flow to $S_d$. 
Due to the resolution mismatch, source images are not available to $S_d$. So the input of $S_d$ is supposed to preserve feature information of anomaly-free data as much as possible.
Referring to \cite{deng2022reverse}, we conduct feature fusion on intermediate outputs of $S_e$ and embed the features into a low-dimensional space, \emph{i.e.} embeddings.
The DFE block has two benefits. First, as described above, DFE can carry rich semantic information which can help $S_d$ comprehend the knowledge of anomaly-free distributions in the training phase.
Second, transferring embeddings instead of original feature maps can enhance the representative diversity on anomaly data in the inference phase.

Fig. \ref{fig4} describes the DFE process in detail. Based on ResNet18, there are three intermediate feature maps involved in the fusion. $F_1$, $F_2$, and $F_3$ represent the feature maps of the three layers with increasing depth, respectively. $F_1$ and $F_2$ are downsampled twice and once by a convolutional blocks with a kernel size of $3\times 3$ and a stride of $2$. 
After be resizing to the same size, the three feature maps are concatenated together.
And then the embedded feature map $F_{emb}$ can be obtained through a residual block \cite{he2016resnet}, in which a convolutional block with a $1\times 1$ kernel is used in the shortcut connection to avoid degradation problem.

\subsection{Anomaly Inference}
In the anomaly inference phase, an input image is judged as anomaly or not according to the anomaly score. Because anomalous patterns at the pixel level mean that defects and faults exist in this image, the detection results at the image level can be obtained from the localization results.

In DSKD, we evaluate the representation discrepancy between $T$ and $S_d$ to score each pixel for anomaly localization, as shown in Fig. \ref{architecture}. Anomaly maps of all deep layers are upsampled to the same size as inputs firstly and then summed up to a map as follows:
\begin{equation}\label{fusion}
\begin{split}
    M(I_m)=\sum_{k=1}^KUpsample(M_{t-d}^k(I_m)),
\end{split}
\end{equation}
where $K$ is the total number of intermediate layers involved. For denoising, $M(I_m)$ is processed by a Gaussian filter with parameter $\sigma=4$ as follows:
\begin{equation}\label{denoise}
\begin{split}
    \widetilde{M}(I_m)=G_{\sigma}(M(I_m)), ~\sigma=4.
\end{split}
\end{equation}
The value of point $(i, j)$ in the final anomaly map $\widetilde{M}(I_m)$ corresponds to the anomaly score on a pixel in an image. And we take the maximum value in $\widetilde{M}(I_m)$ as the anomaly score for the detection result, as shown in Eq. \eqref{score}. And the whole process of anomaly inference is exhibited in Algo. \ref{inference}
\begin{equation}\label{score}
\begin{split}
    score=max(\widetilde{M}(I_m)).
\end{split}
\end{equation}

\begin{table*}\centering \tiny
\caption{Image-level AUROC(100\%) Results on MVTec AD}
\label{table1}
\renewcommand{\arraystretch}{1.25}
\begin{tabular}
{p{35pt}|p{35pt}<{\centering} p{35pt}<{\centering} p{30pt}<{\centering} p{25pt}<{\centering} p{25pt}<{\centering} p{28pt}<{\centering} p{28pt}<{\centering} p{20pt}<{\centering} p{23pt}<{\centering} p{23pt}<{\centering} p{25pt}<{\centering} p{20pt}<{\centering}}
\Xhline{1.2pt}
Method/Category & Patch SVDD\cite{yi2020patch} & AE-SSIM\cite{bergmann2018improving} & AnoGAN\cite{schlegl2017anogan} & Spade\cite{cohen2021spade} & Padim\cite{defard2020padim} & CutPaste\cite{li2021cutpaste} & MB-PFM\cite{wan2022pfm} & DSN\cite{Tao2022dsn} & MKD\cite{salehi2020multiresolution} & STPM\cite{wang2021student} & FAVAE\cite{dehaene2020favae} & Ours \\
\Xhline{0.8pt}
Carpet & 92.9 & 67.0 & 49.0 & 92.8 & 99.8 & 93.9 & \textbf{100} & 96.8 & 79.3 & 98.9 & 67.1 & \textbf{100} \\
Grid & 94.6 & 69.0 & 51.0 & 47.3 & 96.7 & \textbf{100} & 98.0 & 95.6 & 78.0 & \textbf{100} & 97.0 & \textbf{100} \\
Leather & 90.9 & 46.0 & 52.0 & 95.4 & \textbf{100} & \textbf{100} & \textbf{100} & 91.8 & 95.1 & \textbf{100} & 67.5 & \textbf{100} \\
Tile & 97.8 & 52.0 & 51.0 & 96.5 & 98.1 & 94.6 & \textbf{99.6} & 96.4 & 91.6 & 95.5 & 80.5 & 98.4 \\
Wood & 96.5 & 83.0 & 68.0 & 95.8 & 99.2 & 99.1 & 99.5 & 98.3 & 94.3 & 99.2 & 94.8 & \textbf{99.6} \\
Bottle & 98.6 & 88.0 & 69.0 & 97.2 & 99.9 & 98.2 & \textbf{100} & \textbf{100} & 99.4 & \textbf{100} & 99.9 & \textbf{100} \\
Cable & 90.3 & 61.0 & 53.0 & 84.8 & 92.7 & 81.2 & 98.8 & 98.3 & 89.2 & 92.3 & 95.0 & \textbf{99.2} \\
Capsule & 76.7 & 61.0 & 58.0 & 91.0 & 91.3 & \textbf{98.2} & 94.5 & 91.6 & 80.5 & 88.0 & 80.4 & 93.1 \\
Hazelnut & 92.0 & 54.0 & 50.0 & 88.1 & 92.0 & 98.3 & \textbf{100} & 99.4 & 98.4 & \textbf{100} & 99.3 & \textbf{100} \\
Metal nut & 94.0 & 54.0 & 50.0 & 71.0 & 98.7 & 99.9 & 100 & 97.7 & 73.6 & \textbf{100} & 85.2 & 99.7 \\
Pill & 86.1 & 60.0 & 68.0 & 80.1 & 93.3 & 94.9 & \textbf{96.5} & 89.5 & 82.7 & 93.8 & 82.1 & 96.2 \\
Screw & 81.3 & 51.0 & 35.0 & 66.7 & 85.8 & 88.7 & 91.8 & \textbf{98.1} & 83.3 & 88.2 & 83.7 & 97.1 \\
Toothbrush & \textbf{100} & 74.0 & 57.0 & 88.9 & 96.1 & 99.4 & 88.6 & \textbf{100} & 92.2 & 87.8 & 95.8 & 98.9 \\
Transistor & 91.5 & 52.0 & 67.0 & 90.3 & 97.4 & 96.1 & \textbf{97.8} & 91.3 & 85.6 & 93.7 & 97.2 & 97.5 \\
Zipper & 97.9 & 80.0 & 59.0 & 96.6 & 90.3 & \textbf{99.9} & 97.4 & 96.1 & 93.2 & 93.6 & 93.2 & 97.4 \\
\Xhline{0.8pt}
Mean & 92.1 & 63.0 & 55.0 & 85.5 & 95.5 & 96.1 & 97.5 & 96.1 & 87.8 & 95.4 & 87.9 & \textbf{98.5} \\
\Xhline{1.2pt}
\end{tabular}
\end{table*}

\begin{table*}\centering \tiny
\caption{Pixel-level AUROC(100\%)/PRO(100\%) Results in MVTec AD}
\label{table2}
\renewcommand{\arraystretch}{1.25}
\begin{tabular}
{p{35pt}|p{35pt}<{\centering} p{35pt}<{\centering} p{30pt}<{\centering} p{25pt}<{\centering} p{25pt}<{\centering} p{28pt}<{\centering} p{28pt}<{\centering} p{20pt}<{\centering} p{23pt}<{\centering} p{23pt}<{\centering} p{25pt}<{\centering} p{20pt}<{\centering}}
\Xhline{1.2pt}
Method/Category & Patch SVDD\cite{yi2020patch} & AE-SSIM\cite{bergmann2018improving} & AnoGAN\cite{schlegl2017anogan} & Spade\cite{cohen2021spade} & Padim\cite{defard2020padim} & CutPaste\cite{li2021cutpaste} & MB-PFM\cite{wan2022pfm} & DSN\cite{Tao2022dsn} & MKD\cite{salehi2020multiresolution} & STPM\cite{wang2021student} & FAVAE\cite{dehaene2020favae} & Ours \\
\Xhline{0.8pt}
Carpet & 92.6/-- & 87.0/64.7 & 54.0/20.4 & 97.5/94.7 & 99.1/96.2 & 98.3/-- & \textbf{99.2}/96.9 & 99.1/-- & 95.6/-- & 98.8/95.8 & 96.0/-- & \textbf{99.2}/\textbf{97.7} \\
Grid & 96.2/-- & 94.0/84.9 & 58.0/22.6 & 93.7/86.7 & 97.3/94.6 & 97.5/-- & 98.8/96.0 & 98.1/-- & 91.8/-- & 99.0/96.6 & 99.3/-- & \textbf{99.4}/\textbf{97.7} \\
Leather & 97.4/-- & 78.0/56.1 & 64.0/37.8 & 97.6/97.2 & 99.2/97.8 & \textbf{99.5}/-- & 99.4/98.8 & 99.2/-- & 98.1/-- & 99.3/98.0 & 98.1/-- & 99.4/\textbf{99.1} \\
Tile & 91.4/-- & 59.0/17.5 & 50.0/17.7 & 87.4/75.9 & 94.1/86.0 & 90.5/-- & 96.2/88.7 & 90.9/-- & 82.8/-- & \textbf{97.4}/\textbf{92.1} & 71.4/-- & 94.5/88.2 \\
Wood & 90.8/-- & 73.0/60.5 & 62.0/38.6 & 88.5/87.4 & 94.9/91.1 & 95.5/-- & 95.6/92.6 & 94.1/-- & 84.8/-- & \textbf{97.2}/\textbf{93.6} & 89.9/-- & 94.3/91.6 \\
Bottle & 98.1/-- & 93.0/83.4 & 86.0/62.0 & 98.4/95.5 & 98.3/94.8 & 97.6/-- & 98.4/95.4 & 96.4/-- & 96.3/-- & \textbf{98.8}/95.1 & 96.3/-- & 98.7/\textbf{96.4} \\
Cable & 96.8/-- & 82.0/47.8 & 78.0/38.3 & 97.2/90.9 & 96.7/88.8 & 90.0/-- & 96.7/\textbf{94.2} & 97.1/-- & 82.4/-- & 95.5/87.7 & 96.9/-- & \textbf{97.5}/92.5 \\
Capsule & 95.8/-- & 94.0/86.0 & 84.0/30.6 & \textbf{99.0}/93.7 & 98.5/93.5 & 97.4/-- & 98.3/91.7 & 98.3/-- & 95.9/-- & 98.3/92.2 & 97.6/-- & 98.7/\textbf{94.4} \\
Hazelnut & 97.5/-- & 97.0/91.6 & 87.0/69.8 & \textbf{99.1}/95.4 & 98.2/92.6 & 97.3/-- & \textbf{99.1}/\textbf{96.7} & 98.8/-- & 94.6/-- & 98.5/94.3 & 98.7/-- & 98.8/94.6 \\
Metal nut & 98.0/-- & 89.0/60.3 & 76.0/32.0 & 98.1/94.4 & 97.2/85.6 & 93.1/-- & 97.2/\textbf{94.6} & \textbf{98.3}/-- & 86.4/-- & 97.6/94.5 & 96.6/-- & 96.7/89.5 \\
Pill & 95.1/-- & 91.0/83.0 & 87.0/77.6 & 96.5/94.6 & 95.7/92.7 & 95.7/-- & 97.2/96.1 & 96.7/-- & 89.6/-- & \textbf{97.8}/\textbf{96.5} & 95.3/-- & 97.1/94.7 \\
Screw & 95.7/-- & 92.0/88.7 & 80.0/46.6 & 98.9/\textbf{96.0} & 98.5/94.4 & 96.7/-- & 98.7/93.4 & 99.3/-- & 96.0/-- & 98.3/93.0 & 99.3/-- & \textbf{99.4}/94.2 \\
Toothbrush & 98.1/-- & 96.0/78.4 & 90.0/74.9 & 97.9/93.5 & 98.8/93.1 & 98.1/-- & 98.6/90.7 & 98.6/-- & 96.1/-- & 98.9/92.2 & 98.7/-- & \textbf{99.1}/\textbf{94.5} \\
Transistor & 97.0/-- & 90.0/72.5 & 80.0/54.9 & 94.1/\textbf{87.4} & 97.5/84.5 & 93.0/-- & 87.8/74.9 & 87.0/-- & 76.5/-- & 82.5/69.5 & \textbf{98.4}/-- & 91.5/80.2 \\
Zipper & 95.1/-- & 88.0/66.5 & 78.0/46.7 & 96.5/92.6 & 98.5/\textbf{95.9} & \textbf{99.3}/-- & 98.2/94.8 & 98.2/-- & 93.9/-- & 98.5/95.2 & 96.8/-- & 97.5/93.5 \\
\Xhline{0.8pt}
Mean & 95.7/-- & 87.0/69.4 & 74.3/44.3 & 96.5/91.7 & \textbf{97.5}/92.1 & 96.0/-- & 97.3/93.0 & 96.7/-- & 90.7/-- & 97.0/92.1 & 95.3/-- & \textbf{97.6}/\textbf{93.5} \\

\Xhline{1.2pt}
\end{tabular}
\end{table*}

\begin{figure*}[!t]
\centering
\includegraphics[width=2\columnwidth]{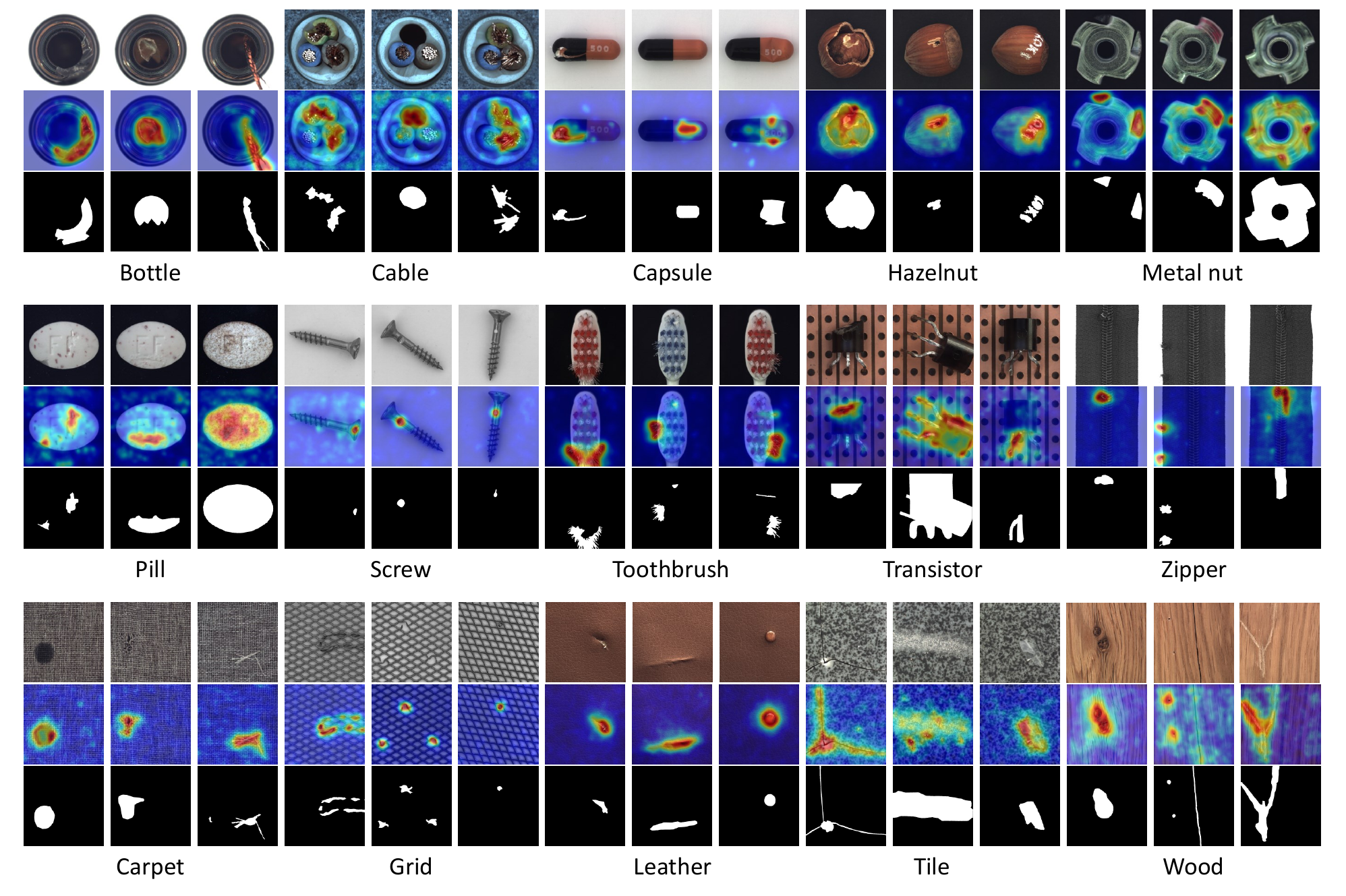}
\caption{Visualization of anomaly localization results on 15 categories in MVTec AD. The first, second and third rows respectively represent the original defective images, anomaly maps and ground truth data}
\label{fig5}
\end{figure*}

\section{Experimental results}
\label{sec:experiments}
In this section, we discuss the experiments and the results. The proposed DSKD model is compared with other methods. In the ablation experiment, we analyze the function of each module in DSKD.

\subsection{Experiment Settings}
\subsubsection{Datasets}
We performed experiments on three datasets, MVTec AD \cite{bergmann2021mvtec}, MVTec 3D-AD \cite{bergmann2021mvtec3d}, Magnetic Tile Defects \cite{huang2020mt}. \textbf{MVTec AD} is a benchmark dataset for unsupervised anomaly detection. It consists of over 5000 high-resolution industrial images, which are divided into 15 categories.
Each category has about 240 anomaly-free data for training and about 100 images for test. And the pixel-level ground truth data can be used to evaluate the results of anomaly segmentation. 
And \textbf{MVTec 3D-AD} is a comprehensive 3D dataset for AD and AL, which contains more than 4,000 high-resolution industrial images in 10 categories. 
Each 3D image has two formats, XYZ for 3D data and RGB for 2D data. In this paper, we only use the RGB images to test our methods.
\textbf{MT Defects} is a dataset for surface defect detection of magnetic tiles, which contains five categories of defects.

\subsubsection{Implement and Environment Details}
All of the networks, including $T$, $S_e$ and $S_d$ are based on ResNet18. 
The input images are firstly resized, $256\times 256$ for MVTec AD and MVTec 3D-AD, and $128\times 128$ for MT Defects.
The resized images are then normalized by $\ell_2$ function. 
In the training phase, we use the adaptive moment estimation algorithm (Adam) as the optimizer, with $\beta=(0.5, 0.999)$. The learning rate is 0.001, and the model is trained in 200 epochs. As for the loss function, $\lambda$ is set to 0.1.
The experiments on three datasets are compared with other approaches. The ablation experiment is used to prove the effectiveness of each module in the dual-student architecture. 

\subsubsection{Evaluation Criteria}
We employ three commonly used evaluation criteria in this paper, including the area under the receiver operating characteristic curve (AUROC) at the image level and the pixel level, and the normalized area under the per-region overlap curve (PRO).
AUROC indicates the probability that the predicted value of a positive sample is larger than that of a negative sample in the random selection.
And PRO evaluates the segmentation results by measuring regions of each size with the same weight and the threshold of false positive should be below 0.3. 
AUROC is possibly impacted by the size of anomaly regions, while PRO is not. 
The image-wise AUROC evaluates the performance in image classification (\emph{i.e.} AD), and pixel-wise AUROC and PRO are used for anomaly segmentation (\emph{i.e.} AL), where higher scores mean better performance.
As for model complexity, we compare DSKD with other methods in inference time (inf. time), floating point operations (FLOPs), the scale of parameters, and memory sizes.

\subsection{Results and Discussion}
\subsubsection{MVTec AD}
We performed experiments on the three datasets and the results are compared with other approaches. 
Table \ref{table1} shows the image-level detection results in MVTec AD.
Our method performs best in many categories, including carpet, grid, leather, wood, bottle, cable, and hazelnut. And our method produces the highest mean score at 98.5\%, exceeding our baseline methods, STPM \cite{wang2021student} and MKD \cite{salehi2020multiresolution} by 2.9\% and 10.5\%, respectively.
Regarding anomaly localization, results of the pixel-level AUROC and PRO are shown in Table \ref{table2}. The proposed DSKD is superior to other methods, with 97.6\% AUROC and 93.5\% PRO scores.
In addition, Padim \cite{defard2020padim} and MB-PFM \cite{wan2022pfm} are also competitive, whose AUROC scores are 97.5\% and 97.3\%, and PRO scores are 92.1\% and 93.0\%, respectively. 
 

The qualitative localization results are visualized in Fig.\ref{fig5}.
DSKD can accurately recognize defective patterns and achieve segmentation results that are very close to the ground truth.
However, the limitation is that our method identifies surrounding anomaly-free pixels as anomalies, especially for the toothbrush, tile and wood.

\begin{figure}[!t]
\centering
\includegraphics[width=\columnwidth]{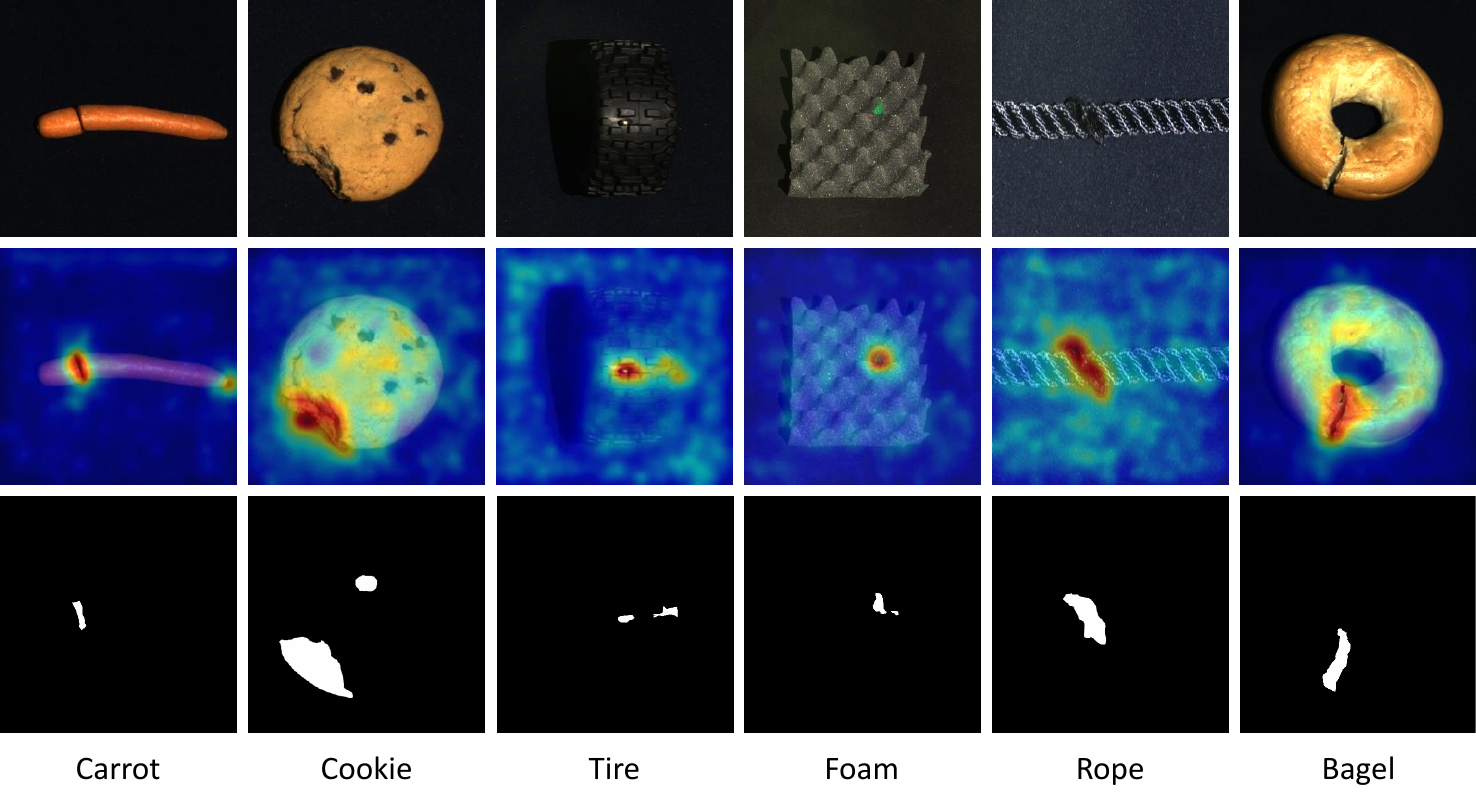}
\caption{Visualization of anomaly localization performance in MVTec 3D-AD.}
\label{fig10}
\end{figure}

\subsubsection{MVTec 3D-AD}
As for MVTec 3D-AD dataset, we conduct experiments on RGB images.
The image-level detection results are shown in Table \ref{table3}. Our method outperforms others a lot.
But the detection AUROC score is only 86.4\%, and DSKD cannot precisely recognize abnormal and normal data in some categories, especially cookies and potato.
However, DSKD performs excellently in anomaly localization, as shown in Table \ref{table4}, where the pixel-level AUROC and PRO are 98.6\% and 95.2\%. 
One possible reason is that the background of an RGB image is usually a uniform color, such as black, which occupies a large percentage of the image and is easy to recognize. 
As a result, the monolithic background can raise the localization score of the entire image. 
And Fig. \ref{fig10} presents the visualized results of pixel-wise localization.
Defective patterns can be accurately recognized, but some neighbor defect-free pixels are sometimes classified as anomalies, \emph{e.g.} the carrot and bagel. And in the cookie sample, the central anomaly is ignored.

\begin{table}\centering \tiny
\caption{Image-level AUROC(100\%) Results in MVTec 3D-AD (RGB)}
\label{table3}
\renewcommand{\arraystretch}{1.25}
\begin{tabular}
{p{35pt}|p{25pt}<{\centering} p{35pt}<{\centering} p{30pt}<{\centering} p{25pt}<{\centering} p{18pt}<{\centering}}
\Xhline{1.2pt}
Method/Category & Padim\cite{defard2020padim} & PatchCore\cite{roth2022patchcore} & CS-flow\cite{rudWeh2022csflow} & STPM\cite{wang2021student} & Ours \\
\Xhline{0.8pt}
Bagel & \textbf{97.5} & 87.6 & 94.1 & 93.0 & 96.1\\
Cable Gland & 77.5 & 88.0 & 93.0 & 84.7 & \textbf{93.0}\\
Carrot & 69.8 & 79.1 & 82.7 & 89.0 & \textbf{90.0}\\
Cookie & 58.2 & 68.2 & \textbf{79.5} & 57.5 & 65.4\\
Dowel & 95.9 & 91.2 & \textbf{99.0} & 94.7 & 97.6\\
Foam & 66.3 & 70.1 & \textbf{88.6} & 76.6 & 84.8\\
Peach & 85.8 & 69.5 & 73.1 & 71.0 & \textbf{88.0}\\
Potato & 53.5 & 61.8 & 47.1 & 59.8 & \textbf{65.5}\\
Rope & 83.2 & 84.1 & \textbf{98.6} & 96.5 & 98.0\\
Tire & 76.0 & 70.2 & 74.5 & 70.1 & \textbf{86.2}\\
\Xhline{0.8pt}
Mean & 76.4 & 77.0 & 83.0 & 79.3 & \textbf{86.4} \\
\Xhline{1.2pt}
\end{tabular}
\end{table}

\begin{table}\centering \footnotesize
\caption{Pixel-level Segmentation Results in MVTec 3D-AD (RGB)}
\label{table4}
\renewcommand{\arraystretch}{1.25}
\begin{tabular}
{p{50pt}|p{50pt}<{\centering} p{50pt}<{\centering}}
\Xhline{1.2pt}
Category & AUROC(100\%) & PRO(100\%)\\
\Xhline{0.8pt}
Bagel & 99.0 & 95.0\\
Cable Gland & 99.1 & 98.8\\
Carrot & 99.2 & 98.5\\
Cookie & 97.9 & 89.0\\
Dowel & 99.3 & 97.6\\
Foam & 95.5 & 85.8\\
Peach & 99.1 & 97.9\\
Potato & 99.0 & 97.6\\
Rope & 99.1 & 94.7\\
Tire & 99.1 & 97.4\\
\Xhline{0.8pt}
Mean & 98.6 & 95.2\\
\Xhline{1.2pt}
\end{tabular}
\end{table}

\subsubsection{MT Defects}
As shown in Table \ref{table5}, the proposed model is dominant in AD with an AUROC score at 96.2\%. But the average results of pixel-level localization are 83.3\% AUROC and 70.3\% PRO.
DSN \cite{Tao2022dsn} achieves the best performance in AL, whose pixel-level AUROC score is at 98.0\%, while our DSKD exceeds DSN by 7.1\% in image-level detection.

\begin{figure}[!t]
\centering
\includegraphics[width=\columnwidth]{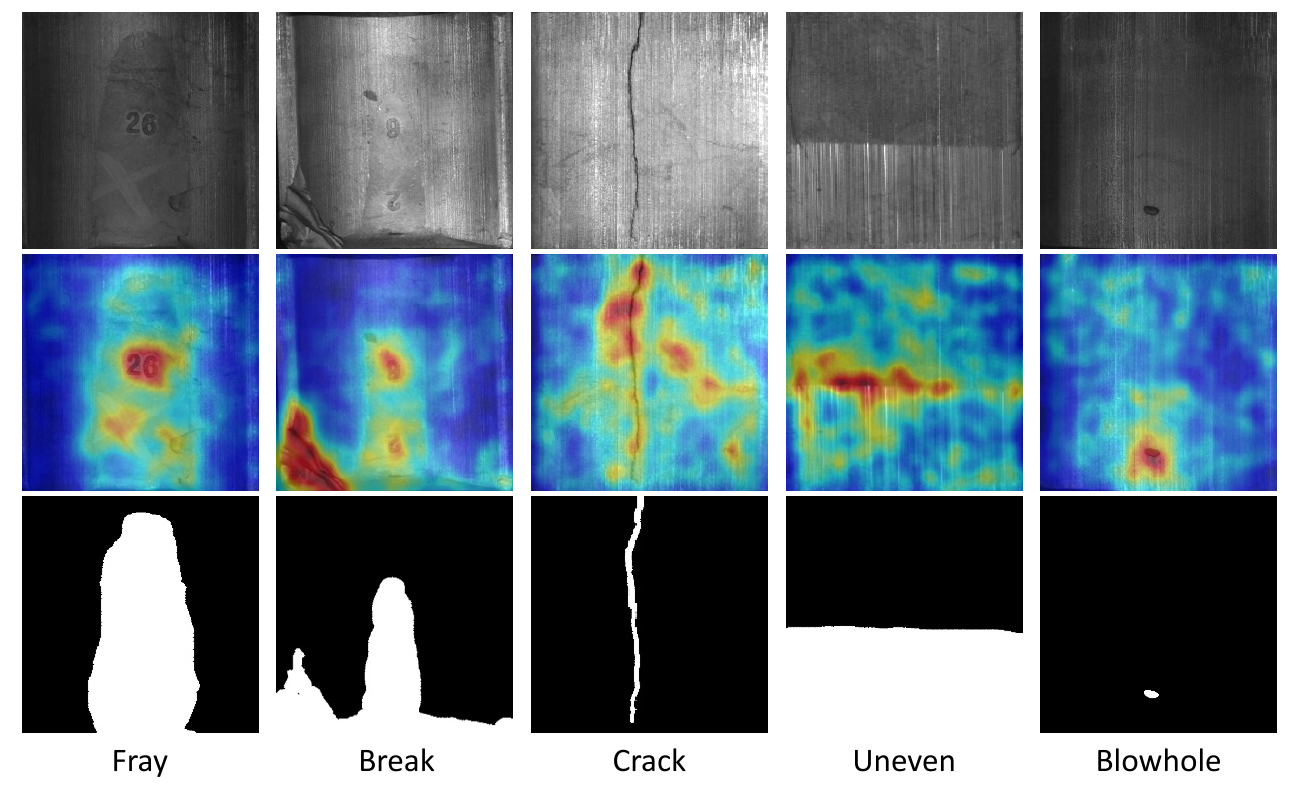}
\caption{Visualization of anomaly localization performance on MT Defect.}
\label{fig6}
\end{figure}

Fig.\ref{fig6} shows the visualization results of AL. There are five defective categories of magnetic tiles including fray, break, crack, uneven, and blowhole. Compared with the ground truth data, our model can recognize most regions with anomalies, which contributes the high accuracy of image-level detection.
Nonetheless, not all defective pixels can be detected with a high anomaly score. For example, in the fray category as shown in Fig.\ref{fig6}, the anomaly region can be highlighted, but many inner pixels are not marked with bright colors, which means that these pixels do not have high enough anomaly scores to be recognized as anomalies. Our model is sensitive to the edge of an anomaly region, whereas inside this area, the data distribution is uniform, which is more like an anomaly-free region. 
Therefore, many pixels cannot be correctly classified, and DSKD gets low accuracy in anomaly localization.

\begin{table}\centering \footnotesize
\caption{Anomaly Detection Results on MT Defects}
\label{table5}
\renewcommand{\arraystretch}{1.25}
\begin{tabular}
{m{50pt}|m{50pt}<{\centering} m{50pt}<{\centering} m{50pt}<{\centering}}
\Xhline{1.2pt}
Method & \makecell[c]{Image-level\\AUROC(100\%)} & \makecell[c]{Pixel-level\\AUROC(100\%)} & PRO(100\%)\\
\Xhline{0.8pt}
AE-SSIM\cite{bergmann2018improving} & 60.1 & 78.3 & -\\
RIAD\cite{zavrtanik2021reconstruction} & 75.2 & 82.7 & -\\
FAVAE\cite{dehaene2020favae} & 83.4 & 86.2 & -\\
DSN\cite{Tao2022dsn} & 89.1 & \textbf{98.0} & -\\
Ours & \textbf{96.2} & 83.3 & \textbf{70.3}\\
\Xhline{1.2pt}
\end{tabular}
\end{table}

\begin{table}\centering \footnotesize
\caption{Comparison for the Results of Model Complexity}
\label{table9}
\renewcommand{\arraystretch}{1.25}
\begin{tabular}
{p{50pt}|p{50pt}<{\centering} p{50pt}<{\centering} p{50pt}<{\centering}}
\Xhline{1.2pt}
Method & Padim\cite{defard2020padim} & MKD\cite{salehi2020multiresolution} & Ours \\
\Xhline{0.8pt}
Inf. time(s) & 0.890 & 0.034 & 0.217\\
FLOPs(G) & 22.85 & 20.96 & 6.17\\
Parameters(M) & 68.9 & 15.0 & 39.3\\
Memory(MB)& 3800 & 4 & 105 \\
\Xhline{1.2pt}
\end{tabular}
\end{table}

\subsubsection{Model Complexity}
The efficiency of anomaly inference plays a significant role in industrial inspection. Unlike the training process, anomaly scores in the inference phase are calculated by the CPU, which is an Intel i7, @2.3GHz in our experiments. As shown in Table \ref{table9}, Padim \cite{defard2020padim} costs most resources in computing and memory. In spite of the high FLOPs at 20.96GB, MKD \cite{salehi2020multiresolution} achieves the best efficiency with 0.034s for inference and takes the smallest scale of parameters and memory. But MKD cannot perform well in AD and AL, as shown in Table \ref{table1} and \ref{table2}. Our method has the fewest FLOPs at 6.17G. And compared with the other methods, DSKD has lower complexity in inference time and memory spaces.

\subsection{Ablation Experiment}
\subsubsection{Influence of Dual-Student Model Architecture}
\begin{figure}[!t]
\centerline{\includegraphics[width=\columnwidth]{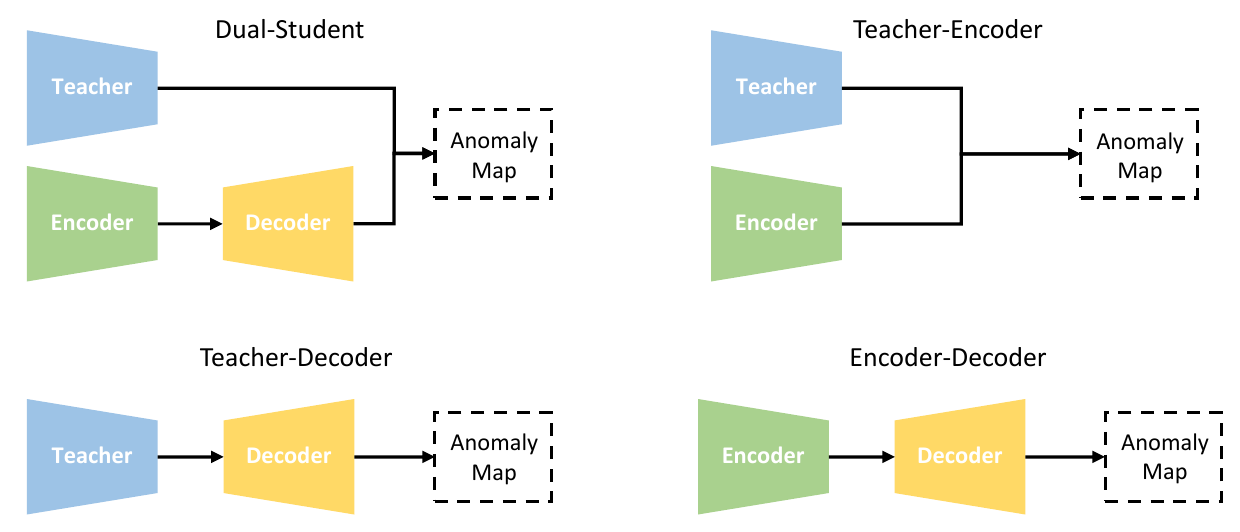}}
\caption{Structures in ablation experiments. Both deep feature embedding and multi-scale feature fusion blocks are used in this experiment.}
\label{fig7}
\end{figure}

\begin{figure*}
    \centering
    \includegraphics[width=2\columnwidth]{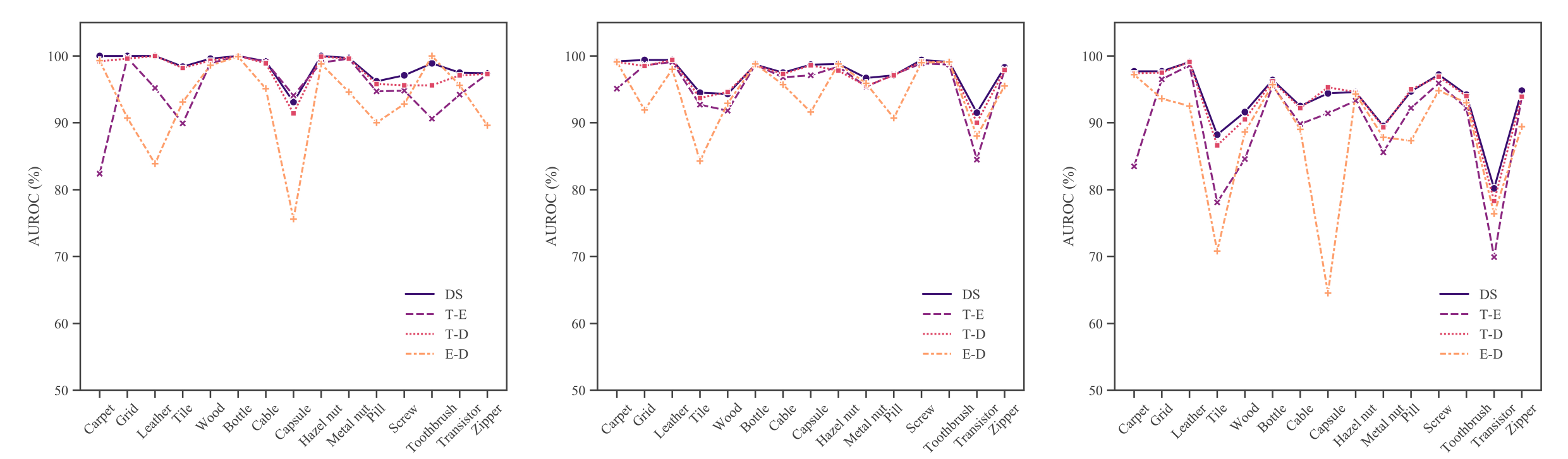}
    \caption{Results for each category in MVTec AD in ablation experiments.}
    \label{fig8}
\end{figure*}

To analyze the effectiveness of the proposed method based on dual-student networks, we conduct experiment on MVTec AD and compare the results with other three models with different structures, as shown in Fig.\ref{fig7}. 
The dual-student, teacher-encoder, teacher-decoder and encoder-decoder structures are respectively marked as DS, T-E, T-D and E-D. Both of the image-level and pixel-level detection results are shown in Table \ref{table6}. And the results for each category are shown in Fig.\ref{fig8}. The DS model is much superior to others. The second best is T-D, with AUROC at the image level and pixel level, and PRO scores at 97.8\%, 97.1\% and 93.1\%, respectively.
Consequently, the discrepancy in the representation of anomalous features between $T$ and $S_d$ is the greatest, while the smallest is between $S_e$ and $S_d$. 
Besides, this experiment can also demonstrate that interaction between $S_e$ and $S_d$ can effectively strengthen the ability to detect the representative discrepancy and improve the performance.

\begin{table}\centering \footnotesize
\caption{Ablation Experiments on Model Architecture on MVTec AD}
\label{table6}
\renewcommand{\arraystretch}{1.25}
\begin{tabular}
{m{35pt}<{\centering}|m{50pt}<{\centering} m{50pt}<{\centering} m{50pt}<{\centering}}
\Xhline{1.2pt}
Structure & \makecell[c]{Image-level\\AUROC(100\%)} & \makecell[c]{Pixel-level\\AUROC(100\%)} & PRO(100\%)\\
\Xhline{0.8pt}
DS & \textbf{98.5} & \textbf{97.6} & \textbf{93.5}\\
T-E & 95.3 & 96.1 & 89.4\\
T-D & 97.8 & 97.1 & 93.1\\
E-D & 93.2 & 94.6 & 87.7\\
\Xhline{1.2pt}
\end{tabular}
\end{table}

\subsubsection{Influence of Deep Feature Embedding}
To study the function of the DFE module, we remove the multiscale feature map fusion step between $S_e$ and $S_d$, and the output of $S_e$ is directly downsampled and then fed to $S_d$.
As shown in Table \ref{table7}, the scores of the three metrics, image-level AUROC, pixel-level AUROC and PRO, of the model with DFE module are much superior to those of the model without DFE module by 22.6\%, 7.9\% and 17.0\%.
Therefore, the embedding of $S_e$ carries a lot of semantic information that can help $S_d$ understand the representation of feature-free. And the detection model can eventually achieve better performance.

\begin{table}\centering \footnotesize
\caption{Ablation Experiments on Deep Feature Embedding}
\label{table7}
\renewcommand{\arraystretch}{1.25}
\begin{tabular}
{m{35pt}<{\centering}|m{50pt}<{\centering} m{50pt}<{\centering} m{50pt}<{\centering}}
\Xhline{1.2pt}
DFE & \makecell[c]{Image-level\\AUROC(100\%)} & \makecell[c]{Pixel-level\\AUROC(100\%)} & PRO(100\%)\\
\Xhline{0.8pt}
\ding{55} & 75.6 & 89.6 & 76.5\\
\ding{51} & \textbf{98.5} & \textbf{97.6} & \textbf{93.5}\\
\Xhline{1.2pt}
\end{tabular}
\end{table}

\subsubsection{Influence of Multi-scale Feature Fusion}
In the anomaly inference phase, DSKD uses feature maps of three different sizes and obtains the anomaly map by multi-scale fusion, as discussed in Sec \ref{sec:proposed method}. To investigate the influence of this operation, we compared the results of models that use anomaly maps of different sizes, as shown in Table \ref{table8}. The $M_1$, $M_2$ and $M_3$ correspond to $M_{t-d}^1$, $M_{t-d}^2$ and $M_{t-d}^3$ in Fig. \ref{architecture}, respectively. And $M_{1-3}$ means that the all of the anomaly maps are used. 
$M_2$ performs best in three single-layer anomaly maps, with scores of image-level AUROC, pixel-level AUROC and PRO at 96.5\%, 96.4\% and 91.0\%. But the results of $M_{1-3}$ achieve the highest scores, which prove the advantage of multi-scale anomaly maps fusion.

\begin{table}\centering \footnotesize
\caption{Ablation Experiments on Multiple Feature Maps Fusion}
\label{table8}
\renewcommand{\arraystretch}{1.25}
\begin{tabular}
{m{35pt}<{\centering}|m{50pt}<{\centering} m{50pt}<{\centering} m{50pt}<{\centering}}
\Xhline{1.2pt}
Used Map & \makecell[c]{Image-level\\AUROC(100\%)} & \makecell[c]{Pixel-level\\AUROC(100\%)} & PRO(100\%)\\
\Xhline{0.8pt}
$M_3$ & 92.7 & 94.2 & 88.7\\
$M_2$ & 96.5 & 96.4 & 91.0\\
$M_1$ & 96.0 & 95.9 & 88.0\\
$M_{1-3}$ & \textbf{98.5} & \textbf{97.5} & \textbf{93.5}\\
\Xhline{1.2pt}
\end{tabular}
\end{table}

\section{Conclusion}
\label{sec:conclusion}
In this paper, we proposed a dual-student KD-based model, DSKD, for unsupervised anomaly detection in industrial inspection.
DSKD consists of a teacher network and two inverted student networks.
The experimental results show that our method achieves great performance with low complexity and a small model size.
And the dual-student architecture and its internal modules are proven to be effective in AD and AL.

However, DSKD is not sensitive to the inner area of anomalies, especially for large-area defects. 
On the other hand, if the target region is small, the neighbor pixels are likely to be recognized as anomalies as well. 
To mitigate these problems, feature representation methods with higher robustness are supposed to be explored. 
And we may improve DSKD by optimizing objective functions, adjusting backbones or employing data augmentation, etc.
Furthermore, we only study the performance of DSKD in 2D images, which can be extended to 3D images in the future, which can better show the structure and details of real-world objects.
3D anomaly detection will be a new research trend.

{\small
\bibliographystyle{ieee_fullname}
\bibliography{egbib}
}

\end{document}